\documentclass[letterpaper]{article} 
\usepackage{aaai25}  
\usepackage{times}  
\usepackage{helvet}  
\usepackage{courier}  
\usepackage[hyphens]{url}  
\usepackage{graphicx} 
\usepackage{natbib}  
\usepackage{caption} 
\usepackage{algorithm}
\usepackage{algorithmic}

\usepackage{newfloat}
\usepackage{listings}
\DeclareCaptionStyle{ruled}{labelfont=normalfont,labelsep=colon,strut=off} 
\floatstyle{ruled}
\newfloat{listing}{tb}{lst}{}
\floatname{listing}{Listing}
\usepackage{graphicx} 
\usepackage{caption} 
\usepackage{booktabs} 
\usepackage{multirow} 
\usepackage{array} 
\usepackage{xcolor} 
\usepackage{colortbl} 

\usepackage{subcaption}
\usepackage{float}

\usepackage{amssymb} 
\usepackage{amsfonts} 

\usepackage{graphicx} 
\usepackage{amsmath} 
\usepackage{booktabs} 
\usepackage{multirow} 
\usepackage{array} 
\usepackage{caption} 
\usepackage{lscape} 
\usepackage{rotating} 

\usepackage{newfloat}
\usepackage{listings}
\DeclareCaptionStyle{ruled}{labelfont=normalfont,labelsep=colon,strut=off} 
\floatstyle{ruled}
\newfloat{listing}{tb}{lst}{}
\floatname{listing}{Listing}
\title{NumbOD: A Spatial-Frequency Fusion Attack Against Object Detectors}

\author{
    Ziqi Zhou\textsuperscript{\rm 1,2,3 $*$},
    Bowen Li\textsuperscript{$\dagger$},
    Yufei Song\textsuperscript{$\dagger$},
    Zhifei Yu\textsuperscript{$\dagger$},
    Shengshan Hu\textsuperscript{\rm 1,2,4,5 $\dagger$},\\
    Wei Wan\textsuperscript{\rm 1,2,4,5 $\dagger$}, 
    Leo Yu Zhang\textsuperscript{$\ddagger$}, 
    Dezhong Yao\textsuperscript{\rm 1,2,3 $*$},
    Hai Jin\textsuperscript{\rm 1,2,3 $*$},
}

\affiliations{
    \textsuperscript{\rm 1}National Engineering Research Center for Big Data Technology and System\\
    \textsuperscript{\rm 2}Services Computing Technology and System Lab
    \textsuperscript{\rm 3}Cluster and Grid Computing Lab\\
    \textsuperscript{\rm 4}Hubei Engineering Research Center on Big Data Security\\
    \textsuperscript{\rm 5}Hubei Key Laboratory of Distributed System Security\\
 $*$ School of Computer Science and Technology, 
Huazhong University of Science and Technology \\
 $\dagger$ School of Cyber Science and Engineering,
Huazhong University of Science and Technology\\
$\ddagger$ 
School of Information and Communication Technology, Griffith University \\

\footnotesize{\texttt{\{zhouziqi,libowen2021,yufei17,yzf,hushengshan,wanwei\_0303,dyao,hjin\}@hust.edu.cn}
}
\\
\footnotesize{\texttt{leo.zhang@griffith.edu.au}}
}

\def\ie{\textit{i.e.}}
\def\eg{\textit{e.g.}}

\usepackage{xcolor}

\usepackage[capitalize]{cleveref}
\crefname{section}{Sec.}{Secs.}
\Crefname{section}{Section}{Sections}
\Crefname{table}{Table}{Tables}
\crefname{table}{Tab.}{Tabs.}

\usepackage{bibentry}

\begin{document}
\maketitle
\begin{abstract}
With the advancement of deep learning, \textit{object detectors} (ODs) with various architectures have achieved significant success in complex scenarios like autonomous driving. 
Previous adversarial attacks against ODs have been focused on designing customized attacks targeting their specific structures (\eg, NMS and RPN), yielding some results but simultaneously constraining their scalability.
Moreover, most efforts against ODs stem from image-level attacks originally designed for classification tasks, resulting in redundant computations and disturbances in object-irrelevant areas (\eg, background). 
Consequently, how to design a model-agnostic efficient attack to comprehensively evaluate the vulnerabilities of ODs remains challenging and unresolved.
In this paper, we propose NumbOD, a brand-new spatial-frequency fusion attack against various ODs, aimed at disrupting object detection within images.
We directly leverage the features output by the OD without relying on its internal structures to craft adversarial examples.
Specifically, we first design a dual-track attack target selection strategy to select high-quality bounding boxes from OD outputs for targeting. Subsequently, we employ directional perturbations to shift and compress  predicted boxes and change classification results to deceive ODs. Additionally, we focus on manipulating the high-frequency components of images to confuse ODs' attention on critical objects, thereby enhancing the attack efficiency.
Our extensive experiments on nine ODs and two datasets show that NumbOD achieves powerful attack performance and high stealthiness.

\begin{links}
    \link{Code}{https://github.com/CGCL-codes/NumbOD}
\end{links}

\end{abstract}
    
\section{Introduction}
The triumphs in deep learning have substantially propelled the development of computer vision tasks, such as traffic sign recognition~\cite{tabernik2019deep}, pedestrian re-identification~\cite{zheng2017person}, and medical image segmentation~\cite{ramesh2021review}. 
Despite its promising prospects, existing researches~\cite{FGSM,DeepFool} have demonstrated the vulnerability of \textit{deep neural networks} (DNNs).
Adversaries can induce model misclassifications with minimal, strategically crafted perturbations, like wrongly identifying an image of a dog as a cat.
Although extensive works~\cite{DeepFool, CW,PGD, MIFGSM, hu2021advhash, zhou2023downstream, li2024transfer, wang2025physical} have thoroughly investigated adversarial attacks on classification, the more challenging task of object detection remains far less explored.

 \begin{figure}[!t]
    \centering
    \includegraphics[scale=0.3]{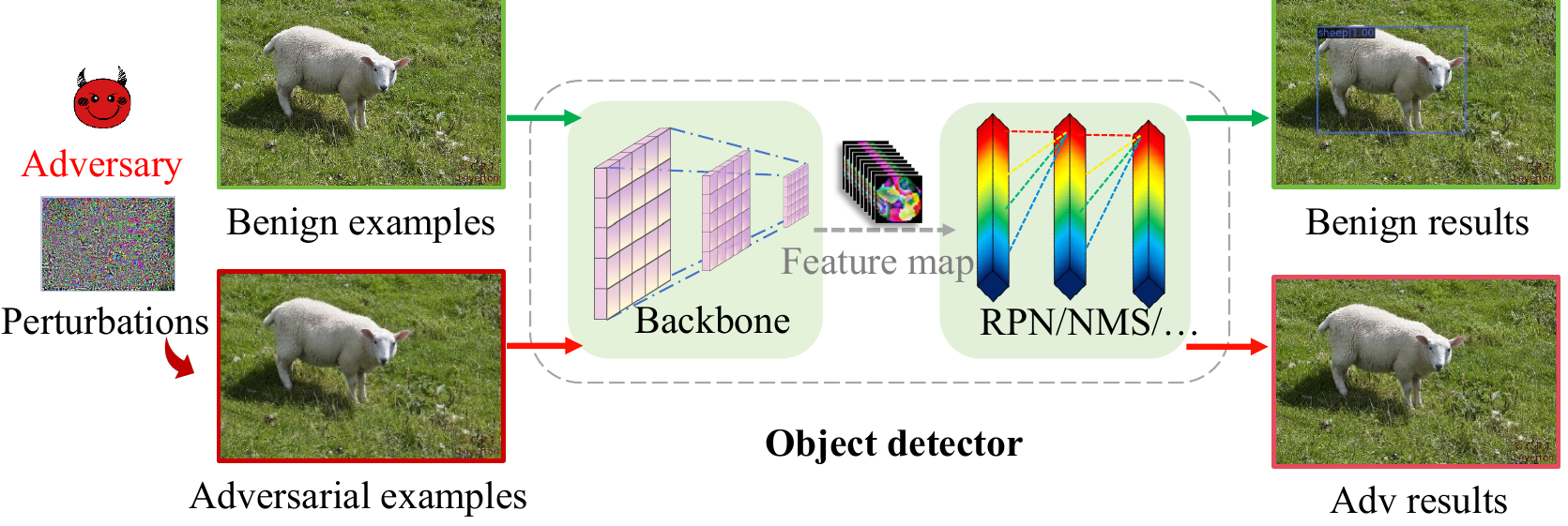}
    \caption{An overview of adversarial examples against an object detector
    }
    \label{fig:demo}
    \vspace{-0.4cm}
\end{figure}

Unlike single classification tasks, object detection involves classification and regression subtasks, requiring simultaneous localization and recognition of objects, \ie, 
providing both bounding boxes and classification results.
Recently, researchers have proposed various modules aimed at enhancing the performance of object detectors, such as \textit{Non-Maximum Suppression} (NMS), \textit{Region of Interest} (RoI) Pooling, and \textit{Region Proposal Networks} (RPN).
The introduction of these new features and modules has brought unprecedented challenges to standard adversarial attacks originally designed for classification tasks.

Recent efforts~\cite{DAG,RAP,CWA} have made decent progress in adversarial attacks against ODs, yet these methods commonly face two major limitations: \textit{1) limited applicability} and \textit{2) low efficiency} in attacks.
Existing efforts have developed effective attacks by exploiting specific vulnerabilities within ODs. 
For example, DAG~\cite{DAG} is the first attack targeting RPN-based models, which implements the attack by minimizing the probability of correct classification. 
RAP~\cite{RAP} enhances the attack by incorporating a loss function tailored for both RPN predicted boxes and classification.
However, due to their reliance on specific modules of ODs, these attack methods greatly limit their scalability, being effective only on detectors with such specific architectural features.
Moreover, existing methods involve image-level global perturbations, which incur unnecessary computational costs by optimizing attacks on non-critical objects, such as the background. 
This may lead to suboptimal attack performance as they simultaneously strive for effective disruption of both meaningful objects and irrelevant background elements.
To the best of our knowledge, how to realize a model-agnostic adversarial attack on critical objects for ODs still remains challenging.

A recent study~\cite{LGP} explored designing adversarial examples without 
$l_{p}$-norm constraints to deceive object detectors with varying architectures. However, to maintain stealthiness, its attack effectiveness is limited, resulting in detection outcomes that still contain some correct bounding boxes.
In contrast, our approach focuses on crafting 
$l_{p}$-norm constrained adversarial attacks that aim to render ODs numb to input images and unable to detect any object, as demonstrated in \cref{fig:demo}. 
In this paper, we propose NumbOD, a novel model-agnostic spatial-frequency fusion attack for ODs.
To achieve a model-agnostic attack, we leverage the final output features of ODs to craft adversarial examples. Our approach employs a dual-track attack target selection strategy, where we independently select the top-k high-quality bounding boxes from both the classification and regression subtasks,
thereby enhancing attack efficiency. 
Upon identifying the attack targets, we design a tailored attack against ODs from both spatial and frequency domains.

Given that object detection involves both classification and regression subtasks, a truly effective attack must simultaneously deceive both components. Specifically, the attack should cause both the predicted bounding boxes and the classification results to deviate from their original outputs.
In the spatial domain, we induce customized deviations in predicted bounding boxes and misclassification results by adding noise to the image. Drawing inspiration from the sensitivity of deep neural networks to frequency components~\cite{luo2022frequency, wang2022invisible}, particularly high-frequency components that capture semantic texture information, we enhance attack efficiency by targeting the image’s high-frequency regions rather than the entire image.
We start by applying the \textit{Discrete Wavelet Transform} (DWT) to decompose the image into high and low-frequency components. We then focus the noise on the semantically significant high-frequency regions, minimizing the discrepancy between adversarial and benign examples and thereby avoiding ineffective attacks on non-critical areas.

We conduct experiments on nine object detectors and two datasets to evaluate the effectiveness of NumbOD.
Both qualitative and quantitative results show that NumbOD effectively deceives ODs of various architectures, exhibiting strong attack performance and high stealthiness. Additionally, comparative experiments reveal that NumbOD surpasses \textit{state-of-the-art} (SOTA) methods for attacking ODs.
Our main contributions are summarized as follows:

\begin{itemize}
\item 
We propose NumbOD, a novel model-agnostic adversarial attack against ODs, designed to disrupt object detection within images across various detector architectures.

\item We design a spatial-frequency fusion attack framework against ODs, which consists of a spatial coordinated deviation attack and a critical frequency interference attack.

\item Our extensive experiments on nine ODs and two datasets show that our NumbOD achieves powerful attack performance and high stealthiness, surpassing SOTA attacks.
 
\end{itemize}
\section{Related works}
\subsection{Object Detectors}
Existing object detection methods are primarily categorized into two distinct paradigms: two-stage detectors~\cite{faster-rcnn,CascadeRCNN,RoITransformer,sabl,xu2020gliding,SparseRCNN,OrientedRCNN,ReDet} and single-stage detectors~\cite{redmon2016you,RetinaNet,reppoints,FCOS,tood,vfnet,S2ANet}.
Two-stage detectors, such as R-CNN~\cite{girshick2014rich}, Faster R-CNN~\cite{faster-rcnn}, and Cascade R-CNN~\cite{CascadeRCNN}, first generate candidate regions through a RPN and then perform precise classification and regression on these regions. 
Conversely, single-stage detectors, such as YOLOs~\cite{redmon2016you,yolov5}, VFNet~\cite{vfnet}, and TOOD~\cite{tood}, 
directly predict object classes and bounding box coordinates across the entire image in a single evaluation step. 
For clarity, we define the objects detected in the images by the object detector as foreground and the remaining parts of the images as background.
Different detectors achieve desirable results in object detection tasks based on their unique modules, which also endow them with distinct vulnerabilities.

\subsection{Adversarial Examples on Object Detectors}
Adversarial example~\cite{FGSM,zhou2023advclip,  zhou2024darksam,zhang2024whydoes,zhou2024securely,song2025segment} is introduced to demonstrate the fragility of DNNs, which involves the addition of minimal perturbations to images, causing misclassification by the target model.
Existing adversarial examples can be categorized into noise-based~\cite{FGSM,DeepFool,PGD} and patch-based methods~\cite{wei2023physically, wei2023unified,huang2023t,qin2023adversarial,tang2023adversarial}. The former boasts high concealment, while the latter offers flexibility but is prone to detection due to its visibility. Therefore, this paper exclusively considers noise-based adversarial methods.

Recently, researchers begin to study the vulnerabilities of object detection, a task that presents greater challenges than classification due to its inclusion of both regression and classification subtasks.
While recent methods~\cite{DAG,RAP,CWA} have demonstrated certain attack performance tailored to specific models, they also inherently limit their scalability, rendering them unsuitable for attacks across different architectural models.
To address this issue, some efforts~\cite{wei2018transferable, TOG,aich2022gama} explored designing preliminary model-agnostic adversarial attacks against ODs.
For example, 
TOG~\cite{TOG} employs two different attack strategies to customize attacks for RPN-based and anchor-based ODs.
However, it cannot attack more recent detectors (\eg, Sparse R-CNN~\cite{SparseRCNN}) without such fundamental components. 
Therefore, there is an urgent need for a truly model-agnostic attack against ODs.
To mitigate adversarial examples, many defenses like data pre-processing,     pruning~\cite{zhu2017prune,ye2018defending}, fine-tuning~\cite{ye2018defending,peng2022fingerprinting}, and adversarial training~\cite{PGD,tramer2019adversarial} have been proposed. These strategies intervene at various stages of model processing to bolster robustness. 
\section{Methodology} \label{sec:method}

\subsection{Problem Formulation}
Object detection is a fundamental task in computer vision, encompassing two subtasks: classification and regression. 
Its output involves providing predicted bounding boxes for target objects, along with corresponding classification labels and scores.
Given an image $x  \in \mathcal{D}$ to an object detector $f(x) \in \mathbb{R}^{N\times (4+1)}$ that returns bounding boxes ${\mathcal{B}}_{n}$ containing the coordinates of the top-left and bottom-right corners
and the  predicted label $Y_{n}, n=1,2,...,N$, with classification score $c_{n} \in \left [ 0,1 \right ] $.

\noindent\textbf{Threat model.} 
We assume that the adversary has access to both the white-box model and the dataset, aiming to design adversarial examples that render the OD ineffective.
Specifically, the adversary aims to craft an elaborate adversarial noise $\delta$ to paste onto the input image $x$ to get an adversarial example $x^{adv}$, which is then fed into the detector to change its original output, \eg, the bounding box is shifted or disappears, the predicted category of the target object changes or the original classification score decreases. 
Note that the noise $\delta$ needs to be small enough to be indistinguishable to the naked eye so that the adversarial examples are not easily detected.
This constraint is typically enforced through an upper bound $\epsilon$ on the $l_{p}$-norm formulated as follow:
\begin{equation} \label{eq:1}
\max_{\delta } \mathbb{E}_{x  \sim  \mathcal{D}}\left [ f\left (  x + \delta  \right )  \ne f\left (  x  \right )  \right ], \quad  s.t.\left \| \delta  \right \| _{p}\le \epsilon
\end{equation}

After feeding the adversarial example $x + \delta$ into the object detector $f(\cdot)$, we can obtain the adversarial prediction boxes ${\mathcal{B}}^{adv}_{n}$, label $Y_{n}^{adv}$ with classification score $c_{n}^{adv}$.

\subsection{Key Challenges and Intuitions}
\begin{figure}[!t]
 \setlength{\abovecaptionskip}{4pt}
    \centering
    \includegraphics[scale=0.34]{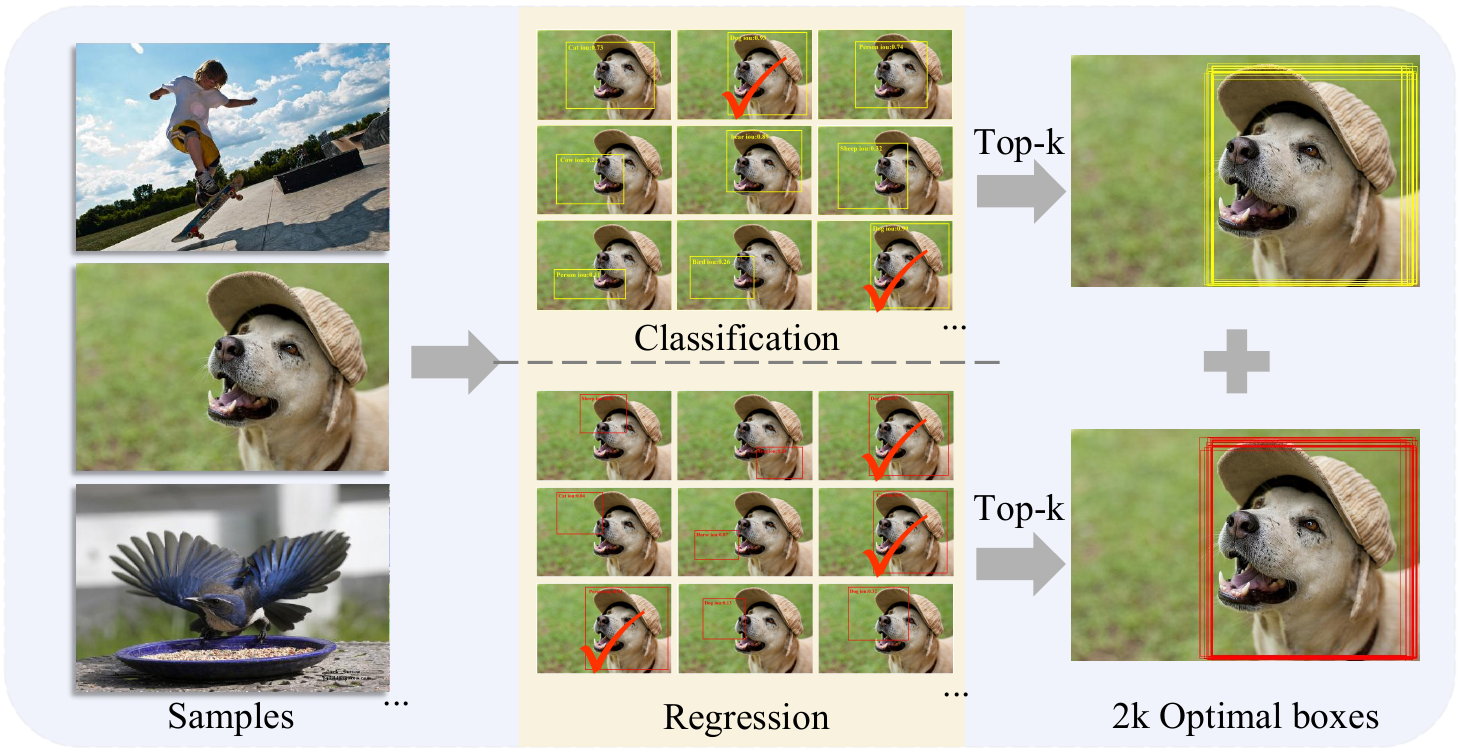}
    \caption{Dual-track attack target selection strategy
    }
    \label{fig:strategy}
    \vspace{-0.4cm}
\end{figure}

Due to the significant structural differences among existing object detectors and their focus on specific object regions within images rather than the entire image, designing a model-agnostic adversarial attack for object detectors presents the following challenges:

\noindent\textbf{Challenge I: The attack dependency on specific modules of object detectors.} 
Benefiting from the designs tailored to the specific modules of object detectors, previous methods have achieved promising attack performance. 
For instance, the customized attack design of RAP~\cite{RAP} focusing on the RPN has proven to be highly effective in deceiving RPN-based detectors. However, this also limits its attack generalization. 
Specifically, RAP demonstrates ineffectiveness in targeting single-stage detectors due to the absence of an RPN structure.
Similarly, other attack methods like FGSM~\cite{FGSM}, DAG~\cite{DAG}, and PGD~\cite{PGD}, which are designed for classification tasks, cannot be directly applied to single-stage models.
Hence, a simple idea is to utilize the final output features of the object detector, which are independent of specific modules, for crafting adversarial examples. However, the extensive bounding boxes generated by the object detector also introduces ambiguity concerning the attack target, thereby incurring unnecessary computational overhead.
Given that object detection algorithms typically employ joint optimization for classification and regression tasks, we propose a \textit{dual-track attack target selection strategy}. 
This strategy enhances attack efficiency by separately selecting the top-k predicted boxes with high scores for both classification and regression tasks as attack targets.
As shown in \cref{fig:strategy}, for regression, we select the predicted boxes with the highest top-k IoU scores for each object in the image as attack targets. For classification, we similarly choose the predicted boxes with the highest top-k IoU scores, but only when the predicted labels match the ground truth labels for each object.
By simultaneously considering the above bounding boxes as attack targets, we aim to enhance attack efficiency while avoiding the emergence of suboptimal attacks.

 \begin{figure*}[!t]
    \centering
    \includegraphics[scale=0.55]{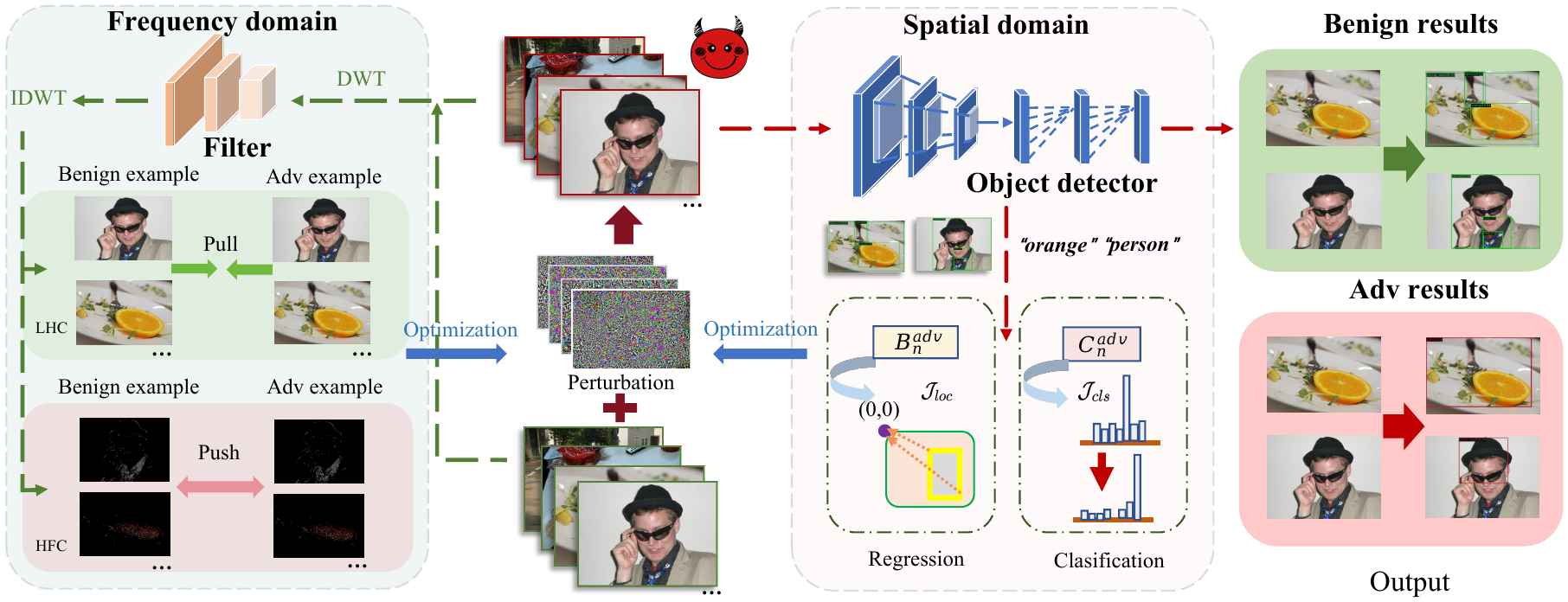}
    \caption{The pipeline of our method
    }
    \label{fig:pipeline}
    \vspace{-0.4cm}
\end{figure*}

\noindent\textbf{Challenge II: The attack redundancy on non-critical objects.} 
Most existing adversarial attacks on object detectors focus on optimizing global noise at the image level. However, perturbing regions outside the target objects (\eg, the background) often fails to enhance attack effectiveness and can lead to inefficiencies.
It is well known that \textit{low-frequency components} (LFC) of an image, which have smooth pixel changes, carry the main information of the image. In contrast, \textit{high-frequency components} (HFC), characterized by abrupt pixel changes, mainly convey details and noise. 
Given that deep neural networks are biased towards image textures, we propose selectively disrupting the HFC of images to hinder the model's recognition of critical objects, thereby increasing the attack's efficiency.
Specifically, we aim to amplify the differences in high-frequency components (\ie, texture information) between adversarial examples and benign samples while constraining the differences in low-frequency components (\ie, shape information). This approach further enhances the attack's effectiveness and stealthiness.
By designing such fusion attack in both spatial and frequency domains, we strategically target crucial areas within images while simultaneously deceiving regression and classification subtasks. 
This provides an efficient optimization direction for generating adversarial examples, resulting in successful attacks on object detectors.

\subsection{Spatial-Frequency Fusion Attack}
In this section, we present NumbOD, a brand-new spatial-frequency fusion attack against object detectors, making them impossible to properly detect objects in images.
The pipeline of NumbOD is depicted in~\cref{fig:pipeline}, which consists of a spatial coordinated deviation attack and a critical frequency interference attack. 
We initially allocate high-quality bounding boxes for each object in the image as attack targets from the perspectives of both regression and classification subtasks, based on the dual-track attack target selection strategy.
Subsequently, in the spatial domain (\ie, traditional attacks involving pixel-level modifications to images), we introduce noise to the images to disrupt the detector's assessment of objects by simultaneously customizing deviations in the positions of predicted boxes and misleading classification outcomes. Simultaneously, in the frequency domain, we enhance the attack performance by further undermining key details, textures, and edges in the images, thereby boosting attack efficiency.
The overall optimization objective of NumbOD is as follow: 
\begin{equation}
\mathcal{J}_{total}= \mathcal{J}_{sa} + \mathcal{J}_{fa}
\label{eq:jt}
\end{equation}
where $\mathcal{J}_{sa}$ is the spatial attack loss and $\mathcal{J}_{fa}$ is the frequency attack loss.

\noindent\textbf{Spatial coordinated deviation attack.} 
Given that the output of the OD for image $x$ primarily includes bounding box locations and classification information. 
Our method targets these two critical components for attack. 
Specifically, we induce a coordinate shift attack ($\mathcal{J}_{loc}$) to alter the size and position of the predicted boxes output by the object detector, and mislead the classification results through a foreground-background separation attack ($\mathcal{J}_{cls}$).
The loss of the spatial coordinated deviation attack is formulated as follow:

\begin{equation}
  \label{eq:st}
\mathcal{J}_{sa} = \mathcal{J}_{loc} + \lambda  \mathcal{J}_{cls}
\end{equation}

For the regression subtask, we design a targeted approach to align the coordinates of the predicted boxes with those of predefined meaningless target regions.
Considering that objects in the image tend to be located in the central area, we force the coordinates of the predicted bounding box's top-left and bottom-right corners to approach the edge point $(0,0)$, causing both positional and size changes to render it ineffective. 
$\mathcal{J}_{loc}$ can be expressed as:

\begin{equation}
\mathcal{J}_{loc} = \sum_{n=1}^{N}  \mathcal{J}_{d}(\mathcal{B}^{adv}_{n},\mathcal{B}^{t}_{n}) / N
\end{equation}
where $\mathcal{B}^{t}_{n}$ represents the target bounding box designed by the attacker, and $\mathcal{J}_{d}$ is the Smooth L$1$ loss.
 
For the classification subtask, we implement a foreground-background separation attack by minimizing the scores associated with the true labels of objects in the image while maximizing the score assigned to the background class. This approach induces the object features within the image to converge towards those of the background, thereby impeding accurate detection.
For the K-class probabilities $c_{n} = (c_{n}^{0},c_{n}^{1},c_{n}^{2},...c_{n}^{K})$, we designate $c_{n}^{gt}$ as the scores attributed to the respective ground truth labels within the $n$-th bounding box.
We enhance the background scores $
c_{n}^{K}$ while reducing the scores $c_{n}^{gt}$ of the corresponding ground truth labels.
The described optimization process can be represented as:

\begin{equation}
\mathcal{J}_{cls} = \sum_{n=1}^{N} \log_{}{(c_{n}^{gt
})} /N - \sum_{n=1}^{N} \log_{}{(c_{n}^{K})}/N
\end{equation}

\begin{table*}[htbp]
  \setlength{\abovecaptionskip}{0.1cm}
  \centering
  \caption{Attack performance of NumbOD against different object detectors}
    \scalebox{0.63}{
    \begin{tabular}{ccccccccccccccccccccc}
    \toprule
    \toprule
    \multicolumn{2}{c}{Datasets} & \multicolumn{9}{c}{MS-COCO}                                           &       & \multicolumn{9}{c}{PASCAL VOC} \\
    \multicolumn{2}{c}{Models} & FR    & CR    & SR    & SFR   &  RP   & VFNet & TOOD  & D.DETR & YOLO &       & FR    & CR    & SR    & SFR   &  RP   & VFNet & TOOD  & D.DETR & YOLO \\
\cmidrule{3-11}\cmidrule{13-21}    \multicolumn{2}{c}{IW-SSIM↓} & 0.17  & 0.18  & 0.12  & 0.17  & 0.15  & 0.16  & 0.13  & 0.18  & 0.17  &       & 0.20  & 0.20  & 0.14  & 0.20  & 0.17  & 0.17  & 0.15  & 0.17  & 0.17 \\
    \multicolumn{2}{c}{NMSE↓} & 0.01  & 0.01  & 0.01  & 0.01  & 0.01  & 0.01  & 0.01  & 0.01  & 0.01  &       & 0.01  & 0.01  & 0.01  & 0.01  & 0.01  & 0.01  & 0.01  & 0.02  & 0.01 \\
    \multicolumn{2}{c}{TV↓} & 96.14 & 96.17 & 96.08 & 96.15 & 96.10 & 96.15 & 96.06 & 96.28 & 96.23 &       & 81.12 & 81.13 & 81.04 & 81.14 & 81.04 & 81.19 & 81.02 & 81.28 & 81.18 \\
    \multirow{2}[0]{*}{mAP50(\%)} & clean↑ & 50.98  & 51.28 & 47.58 & 50.70 & 48.99 & 51.31 & 51.80 & 60.79 & 53.32 &       & 74.45 & 75.09 & 70.60 & 73.83 & 73.68 & 73.95 & 57.41 & 78.51 & 69.42 \\
          & adv↓  & 0.38  & 0.27  & 3.62  & 0.47  & 2.25  & 5.49  & 2.69  & 1.69  & 0.59  &       & 0.54  & 0.22  & 3.21  & 0.71  & 1.35  & 1.96  & 2.54  & 3.22  & 2.14 \\
    \multirow{2}[1]{*}{mAP75(\%)} & clean↑ & 34.74 & 37.55 & 32.54 & 36.93 & 32.89 & 37.68 & 38.86 & 43.73 & 36.84 &       & 57.08 & 60.04 & 52.56 & 58.34 & 56.19 & 59.74 & 41.86 & 62.19 & 51.47 \\
          & adv↓  & 0.06  & 0.08  & 1.17  & 0.10  & 0.74  & 1.90  & 1.32  & 1.32  & 0.17  &       & 0.04  & 0.02  & 1.15  & 0.08  & 0.21  & 0.39  & 0.99  & 2.03  & 0.86 \\
    \bottomrule
    \bottomrule
    \end{tabular}%
    }
   \label{tab:attack_performance}%
   \vspace{-0.2cm}
\end{table*}%

  \begin{figure*}[!t]
 \setlength{\abovecaptionskip}{4pt}
    \centering
    \includegraphics[scale=0.52]{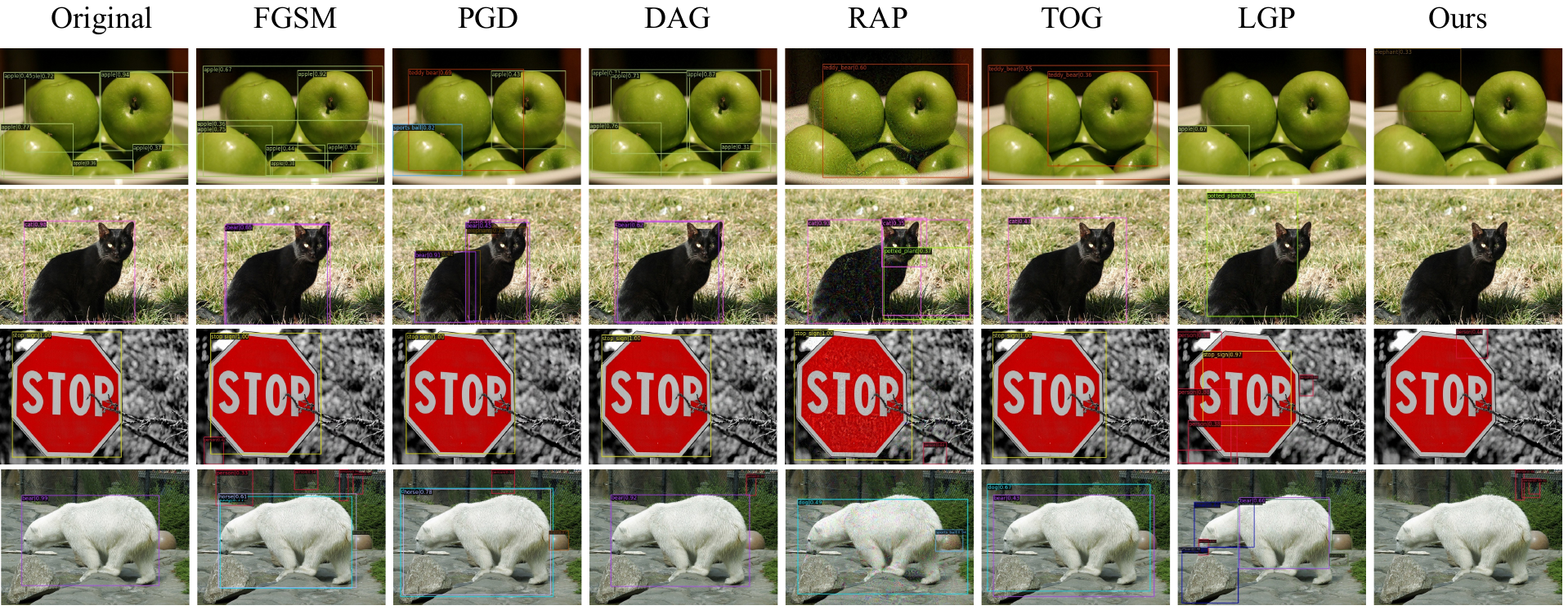}
    \caption{Visualizations of the adversarial examples made by different methods against Faster R-CNN on MS-COCO
    }
    \label{fig:compare}
     \vspace{-0.4cm}
\end{figure*}

\noindent\textbf{Critical frequency interference attack.}  
In the frequency domain, the high-frequency components  of an image denote the finer details, including noise and textures,  while the low-frequency components contain the general outline and overall structural information of the image.
We aim to disrupt the sensitive high-frequency components of DNNs to interfere with the OD's focus on crucial objects in input images.
We employ the DWT, utilizing  a low-pass filter $\mathcal{L}$ and a high-pass filter $\mathcal{H}$,  to decompose the image $x$ into different components, constituting a low-frequency component $c_{ll}$, a high-frequency component $c_{hh}$, and two mid-frequency components $c_{lh}$ and $c_{hl}$, via
\begin{equation}
  \label{eq:dwt}
  c_{ll}   =  \mathcal{L} x  \mathcal{L}^T,   c_{hh}   =  \mathcal{H} x  \mathcal{H}^T, 
   c_{lh} / c_{hl}  =  \mathcal{L} x  \mathcal{H}^T / \mathcal{H} x  \mathcal{L} ^T
\end{equation}

Subsequently, we employ the \textit{inverse discrete wavelet transform} (IDWT) to reconstruct the signal that has been decomposed through DWT into an image.
We choose the LFC and HFC while dropping the other components to obtain the reconstructed images $\phi (x) $ and $\psi (x)$ as
\begin{equation}
  \label{eq:idwtl}
  \phi (x)   =  \mathcal{L} ^T x_{ll}\mathcal{L}  = \mathcal{L} ^T (\mathcal{L}  x \mathcal{L} ^T) \mathcal{L} 
\end{equation}

\begin{equation}
  \label{eq:idwth}
  \psi  (x)   =  \mathcal{H} ^T x_{hh}\mathcal{H}  = \mathcal{H} ^T (\mathcal{H}  x \mathcal{H} ^T) \mathcal{H}
\end{equation}

By adding the adversarial noises to the images, we alter their high-frequency components, disrupting the original texture information. Simultaneously, we enforce constraints on the low-frequency disparities between adversarial and benign examples to redirect a larger portion of the perturbation towards the high-frequency domain, thereby enhancing the attack performance and stealthiness of adversarial examples. 
The loss of the critical frequency interference attack can be expressed as:
\begin{equation} 
\begin{aligned}
\mathcal{J}_{fa} &=  \mathcal{J}_{lfc} -  \mathcal{J}_{hfc} \\ 
&=  \mathcal{J}_{d} (\phi (x),  \phi (x+\delta)) -  \mathcal{J}_{d} (\psi (x), \psi (x+\delta))
\label{eq:ft}
\end{aligned}
\end{equation}

\section{Experiments}
\subsection{Experimental Setup}

\noindent\textbf{Datasets and models.} 
For a comprehensive evaluation, we use MS-COCO~\cite{COCO} and PASCAL VOC~\cite{voc} datasets with victim models
with ResNet50, ResNet101, and ResNeXt101 as backbones.
Unless otherwise specified, we use ResNet50 as the default backbone for evaluation.
Specifically, we select the following nine models: (1) Two-stage detectors: \textit{Faster R-CNN} (FR)~\cite{faster-rcnn}, \textit{Cascade R-CNN} (CR)~\cite{CascadeRCNN}, \textit{SABL Faster R-CNN} (SFR)~\cite{sabl}, and \textit{Sparse R-CNN} (SR)~\cite{SparseRCNN}. 
(2) Single-stage detectors: \textit{RepPoints} (RP)~\cite{reppoints}, \textit{Deformable DETR} (D.DETR)~\cite{DeformableDetr},  VFNet~\cite{vfnet}, TOOD~\cite{tood},
and \textit{YOLO v5} (YOLO)~\cite{yolov5}.

\noindent\textbf{Evaluation metrics.}   
In terms of attack effectiveness, we evaluate the attack performance of our method using the widely adopted metric in the object detection domain, \textit{Mean Average Precision} (mAP). 
We choose mAP$_{50}$ and mAP$_{75}$ as indicators, which  represent the average precision at \textit{Intersection over Union} (IoU) thresholds of $0.5$ and $0.75$, respectively.
In terms of attack stealthiness, we select commonly used metrics such as \textit{Inception Weighted Structural Similarity Index Metric} (IW-SSIM), \textit{Normalized Mean Squared Error} (NMSE), and \textit{Total Variation} (TV) to assess the distance between benign images and perturbed images.
For clarity, we default to multiplying the values of  mAP, IW-SSIM, and NMSE by $100$.

\noindent\textbf{Implementation details.}
Following~\cite{FGSM,DAG,PGD,TOG}, we set the upper bound of the adversarial perturbation to $8/255$.
We set the hyperparameters $\lambda$ to  $100$, while the training epoch is set to $50$ with a batch size of $1$.
We utilize the Adamax optimizer and set the learning rate and weight decay to $0.03$ and $0.02$, respectively.

\begin{table*}[htbp]
\setlength{\abovecaptionskip}{4pt}
  \centering
  \caption{Comparison Study. Bolded values indicate the best results.}
  \scalebox{0.78}{
    \begin{tabular}{ccccccccccccccccc}
    \toprule
    \toprule
    \multirow{2}[4]{*}{Metric} & \multirow{2}[4]{*}{Model} & \multicolumn{7}{c}{MS-COCO}                           &       & \multicolumn{7}{c}{PASCAL VOC} \\
\cmidrule{3-9}\cmidrule{11-17}          &       & FGSM  & PGD   & DAG   & RAP   & TOG   & LGP   & Ours  &       & FGSM  & PGD   & DAG   & RAP   & TOG   & LGP   & Ours \\
    \midrule
    Epsilon & \multirow{6}[2]{*}{\begin{sideways} Faster R-CNN\end{sideways}} & 8     & 8     & 8     & -     & 8     & -     & 8     &       & 8     & 8     & 8     & -     & 8     & -     & 8 \\
    IW-SSIM↓ &       & \textbf{0.16 } & 0.18  & 0.18  & 5.38  & 0.25  & 0.52  & 0.17  &       & 0.20 & 0.21  & \textbf{0.19}  & 4.00  & 10.61  & 0.21  & 0.20 \\
    NMSE↓ &       & \textbf{0.01 } & 0.02  & \textbf{0.01 } & 0.58  & 0.02  & 0.04  & \textbf{0.01 } &       & 0.02  & 0.02  & 0.02  & 0.62  & 1.34  & 0.02  & \textbf{0.01 } \\
    TV↓   &       & 96.92  & 97.24  & 96.28  & 109.97  & 96.40  & 97.04  & \textbf{96.14 } &       & 82.06  & 81.51  & 81.21  & 92.77  & 107.08  & 81.26  & \textbf{81.12 } \\
    mAP50↓ &       & 17.81  & 3.96  & 3.36  & 11.30  & 8.90  & 1.49  & \textbf{0.38 } &       & 33.15  & 7.46  & 5.44  & 47.46  & 5.25  & 3.41  & \textbf{0.54 } \\
    mAP75↓ &       & 8.75  & 1.36  & 1.42  & 4.90  & 3.90  & 0.12  & \textbf{0.06 } &       & 20.08  & 1.90  & 1.82  & 30.98  & 3.23  & 0.42  & \textbf{0.04 } \\
    \midrule
    Metric & \multirow{7}[1]{*}{\begin{sideways}VFNet\end{sideways}} & FGSM  & PGD   & DAG   & RAP   & TOG   & LGP   & Ours &       & FGSM  & PGD   & DAG   & RAP   & TOG   & LGP   & Ours \\
    Epsilon &       & -     & -     & -     & -     & 8     & -     & 8     &       & -     & -     & -     & -     & 8     & -     & 8 \\
    IW-SSIM↓ &       & -     & -     & -     & -     & 0.22  & 0.39  & \textbf{0.16} &       & -     & -     & -     & -     & 0.21  & 0.61  & \textbf{0.17} \\
    NMSE↓ &       & -     & -     & -     & -     & \textbf{0.01} & 0.02  & \textbf{0.01} &       & -     & -     & -     & -     & 0.02  & 0.04  & \textbf{0.01} \\
    TV↓   &       & -     & -     & -     & -     & 96.37 & 96.59  & \textbf{96.15} &       & -     & -     & -     & -     & 81.37 & 81.76  & \textbf{81.19} \\
    mAP50↓ &       & -     & -     & -     & -     & 12.85 & 13.49  & \textbf{5.49} &       & -     & -     & -     & -     & 10.41 & 10.99  & \textbf{1.96} \\
    mAP75↓ &       & -     & -     & -     & -     & 3.92  & 3.49  & \textbf{1.90} &       & -     & -     & -     & -     & 3.82  & 1.57  & \textbf{0.39} \\
        \bottomrule
        \bottomrule
    \end{tabular}%
    }
  \label{tab:compare}%
   \vspace{-0.4cm}
\end{table*}%

\subsection{Attack Performance}
To comprehensively evaluate NumbOD's effectiveness, we conduct experiments on nine object detectors,
with ResNet50 as the backbone, across two datasets, MS-COCO and PASCAL VOC.
For a single attack, we randomly select 5000 images from the dataset to craft adversarial examples, which are then fed into the object detector to evaluate the effectiveness and stealthiness of our approach. 

We first provide a quantitative evaluation of NumbOD in \cref{tab:attack_performance}.
The results reveal the substantial impact of adversarial attacks on the performance of various object detection models across different datasets. Our proposed NumbOD causes significant reductions in mAP scores at both $50\%$ and $75\%$ IoU thresholds, indicating a marked decrease in detection accuracy. Notably, models like Cascade R-CNN and RepPoints show increased vulnerability, with mAP values significantly dropping across the datasets.
We also present the qualitative evaluation results in \cref{fig:compare} to further validate the effectiveness of our approach.
The results in the last column of \cref{fig:compare}  indicate that the object detector fails to detect objects in the adversarial examples generated by NumbOD, with prediction box positions completely wrong or missing, and misclassification results.

Notably, as indicated by the stealthiness metrics in \cref{tab:attack_performance} and the images shown in \cref{fig:compare}, our approach exhibits remarkably high stealthiness. 
The generated adversarial examples are visually indistinguishable, excelling in both visual appearance and stealthiness metrics.
Both qualitative and quantitative experimental results demonstrate the effectiveness and stealthiness of our method in fooling various object detectors, showcasing its high efficacy and stealthiness.

\begin{figure}[!t]  
\setlength{\abovecaptionskip}{4pt}
  \centering
        \subcaptionbox{Module}{\includegraphics[width=0.22\textwidth]{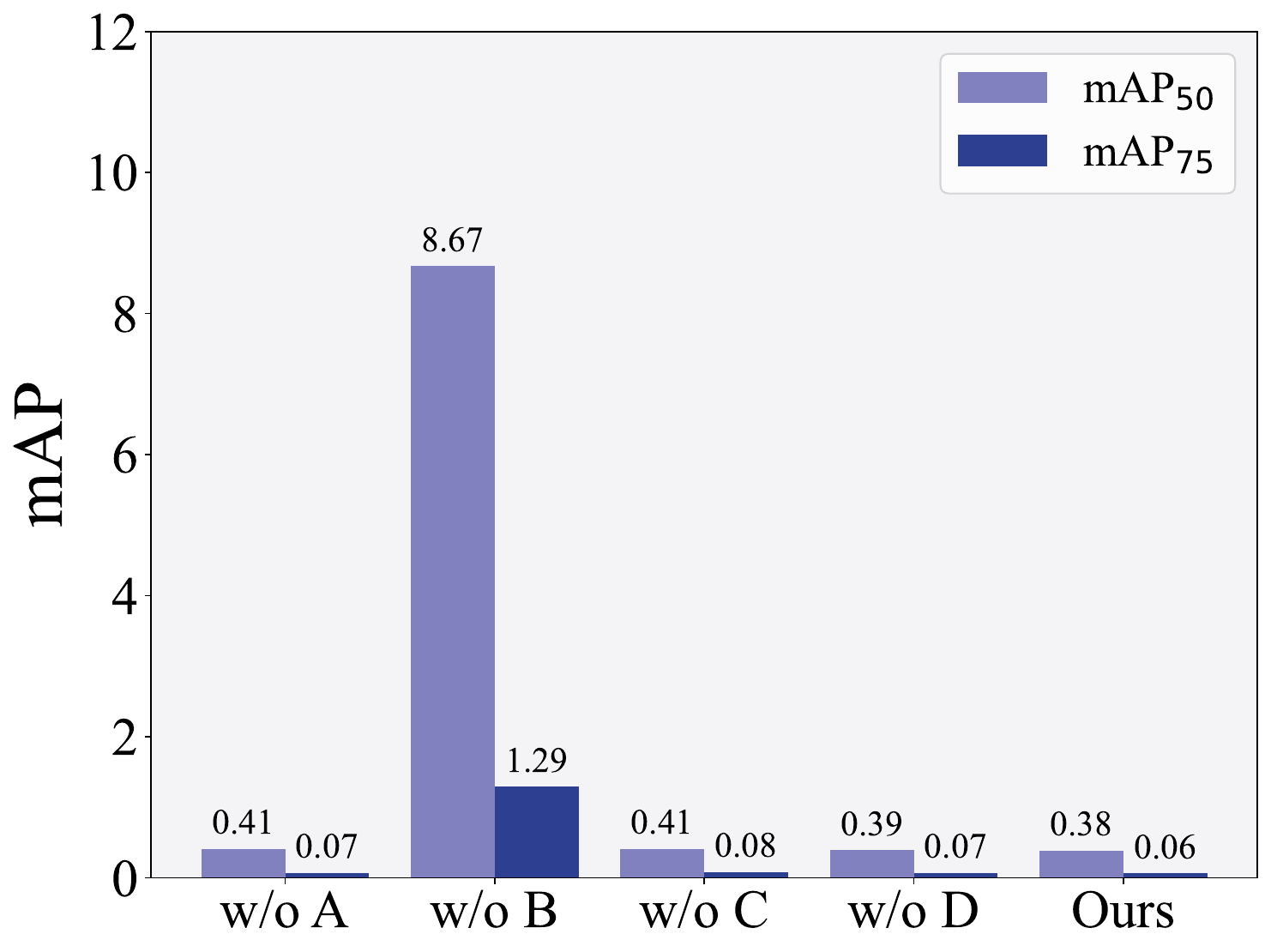}}
    \subcaptionbox{Backbone}{\includegraphics[width=0.22\textwidth]{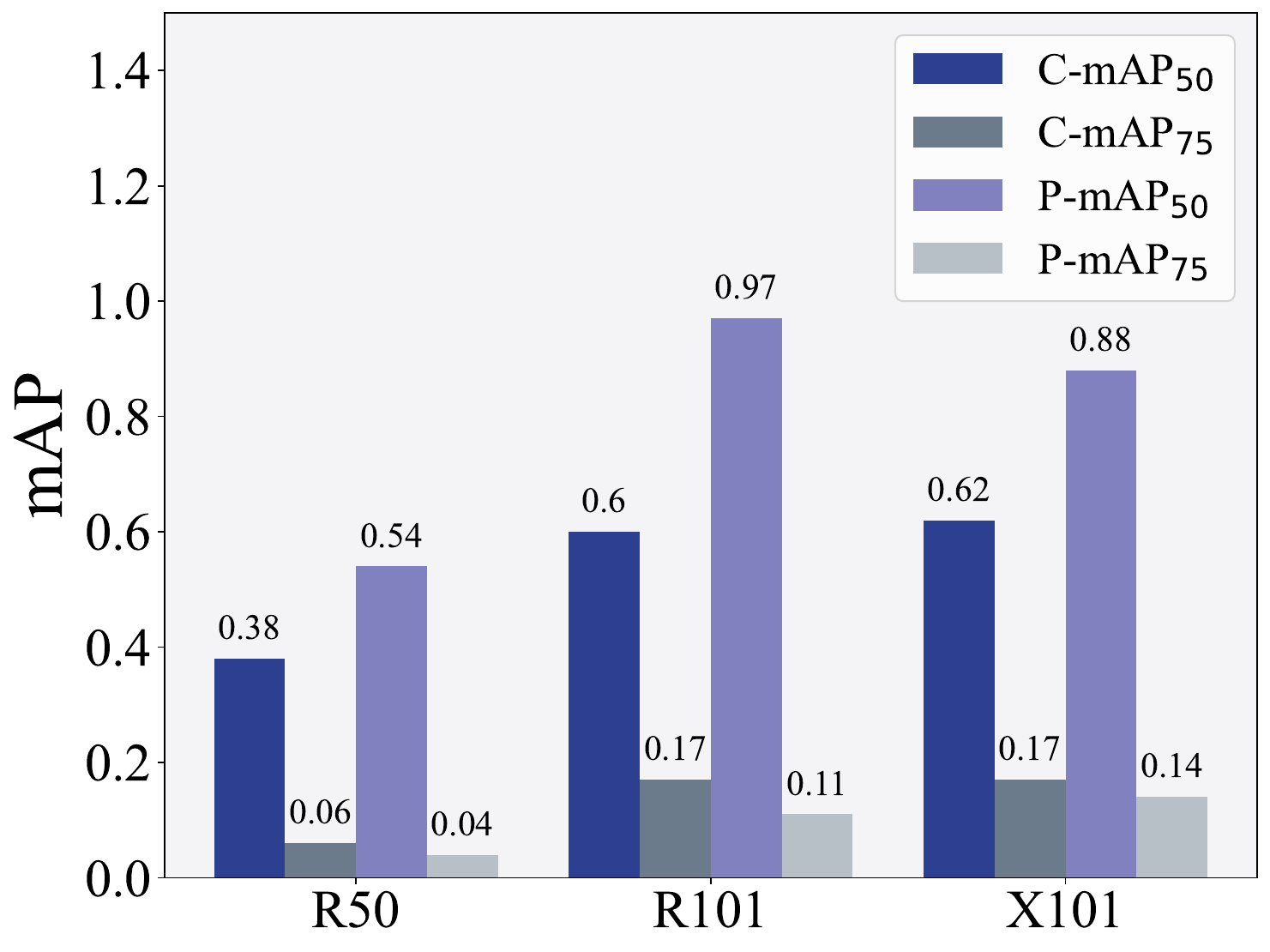}}
      \caption{Ablation Study. C-mAP$_{50}$ and P-mAP$_{50}$ denote the mAP$_{50}$ results on MS-COCO and PASCAL VOC, Others stand the same meaning.}
       \label{fig:ablation_results}
        \vspace{-0.4cm}
\end{figure}

\begin{figure*}[!t]  
\setlength{\abovecaptionskip}{4pt}
  \centering
      \subcaptionbox{Corruption}{\includegraphics[width=0.24\textwidth]{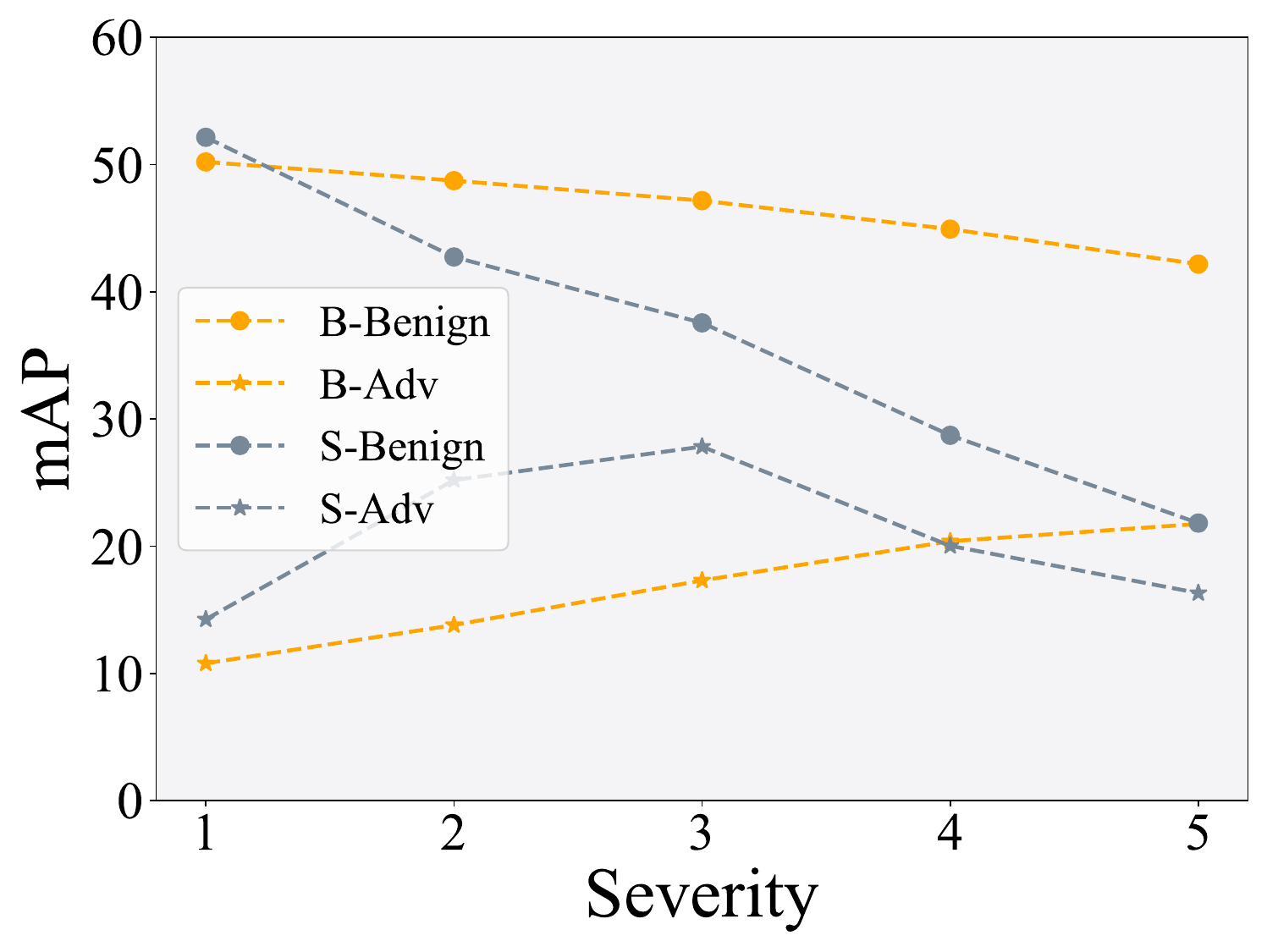}}
        \subcaptionbox{Pruning}{\includegraphics[width=0.24\textwidth]{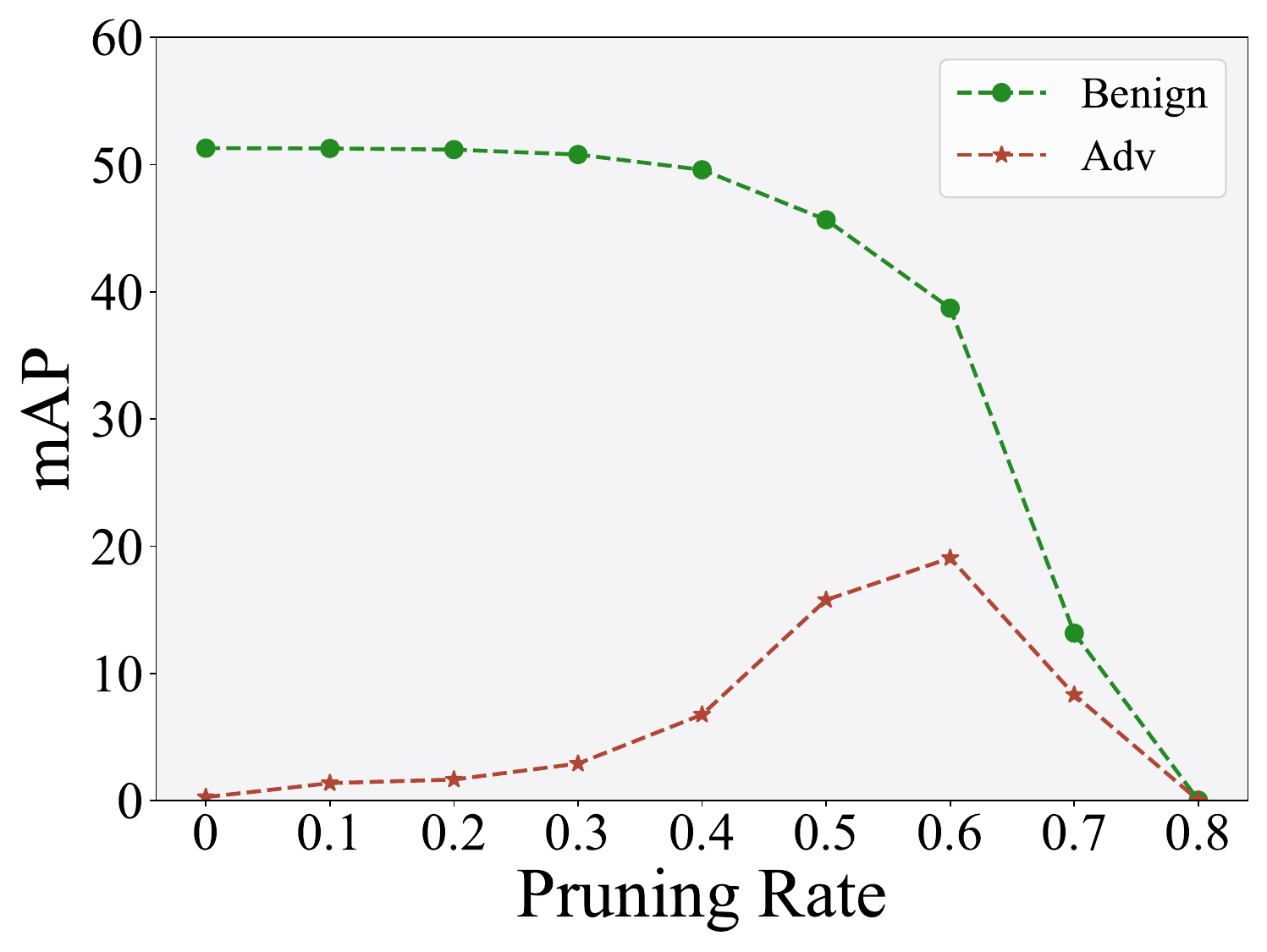}}
    \subcaptionbox{Fine-tuning}{\includegraphics[width=0.2415\textwidth]{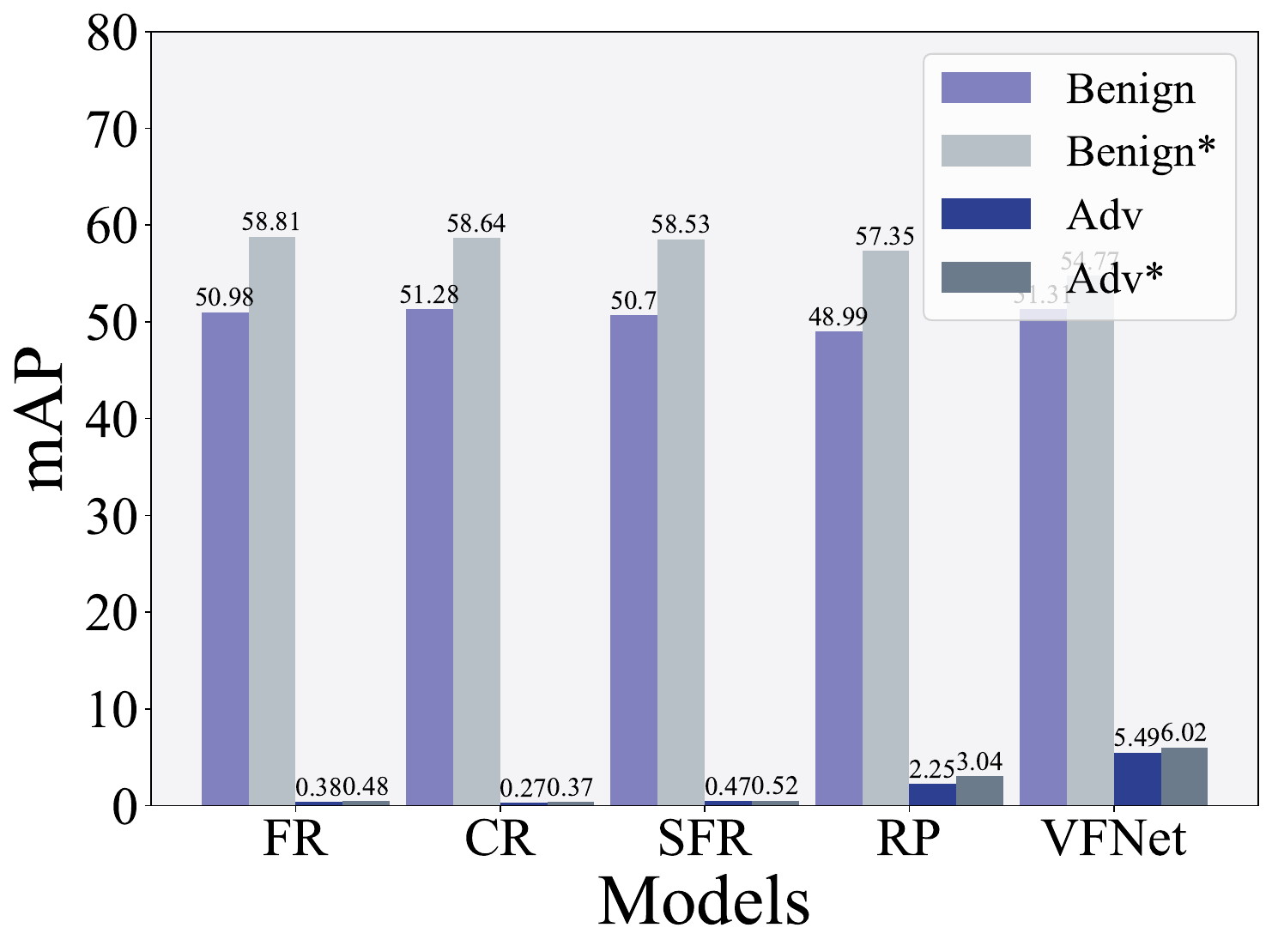}}
      \subcaptionbox{Adversarial Training}{\includegraphics[width=0.2415\textwidth]{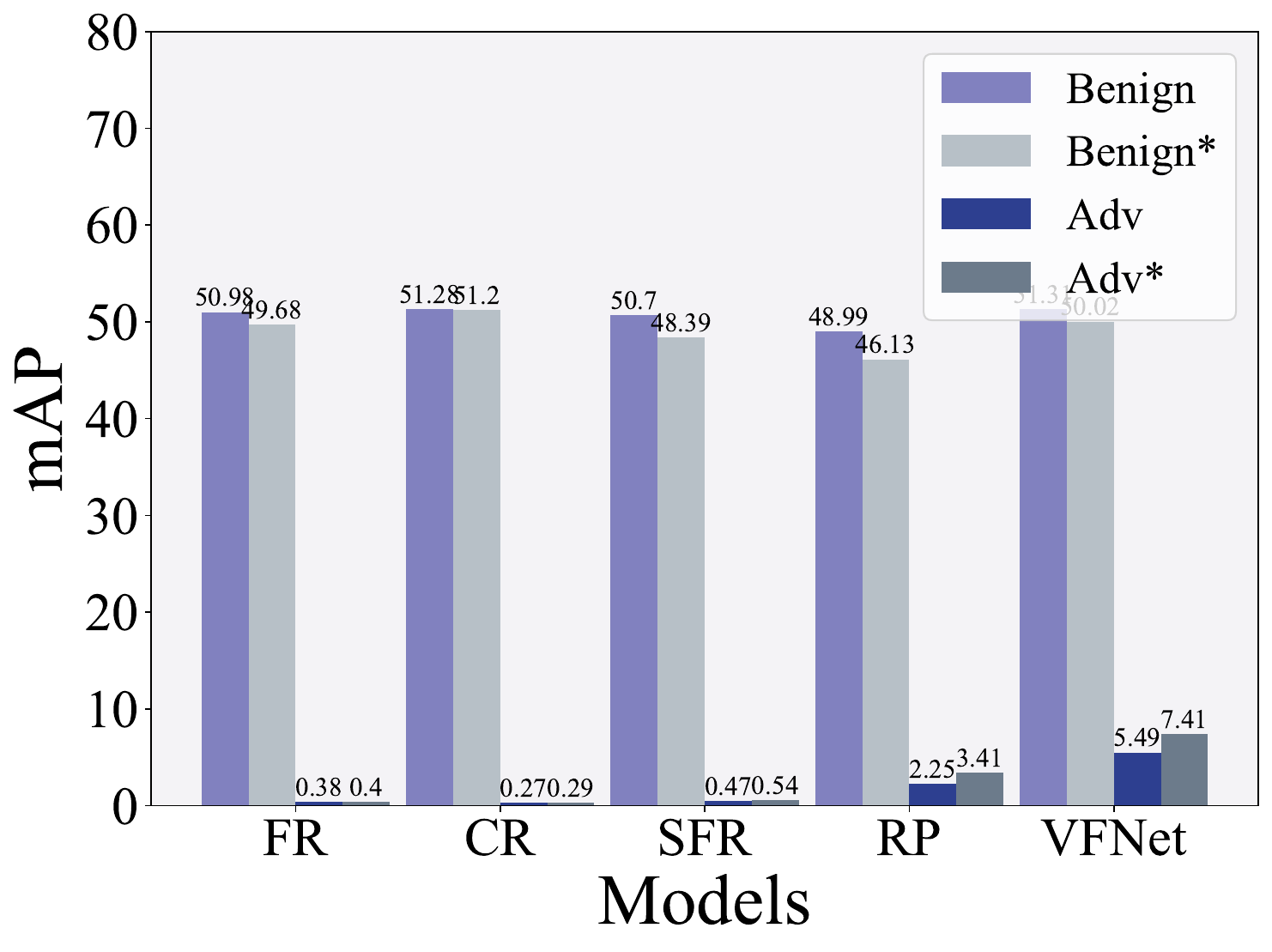}}

      \caption{The attack performance of NumbOD against different defenses on the MS-COCO dataset. (a) - (d) examine four defenses Corruption, Pruning, Fine-tuning, and Adversarial training on our method, respectively. ``Benign$^{*}$'' and ``Adv$^{*}$'' represent the results of the object detector after employing defense methods.}
       \label{fig:defense}
        \vspace{-0.6cm}
\end{figure*}

\subsection{Comparison Study}
To showcase the superiority of our method, we conduct comparative experiments against SOTA adversarial example methods from both effectiveness and stealthiness perspectives.
Specifically, we compare our proposed NumbOD with six popular attack methods, FGSM~\cite{FGSM}, DAG~\cite{DAG}, PGD~\cite{PGD}, RAP~\cite{RAP}, TOG~\cite{TOG}, and LGP~\cite{LGP}, on two models across two datasets.
Among them, LGP is the latest SOTA attack tailored for object detectors.
The perturbation constraints of LGP and RAP do not belong to the $l_{p}$ norm. We use the parameters as stated in their original papers. 
The perturbation budget for the other attacks is set to $8/255$.

We first present the quantitative comparison of our method with these  methods in \cref{tab:compare}. 
The results indicate that our method outperforms all existing approaches in terms of effectiveness and stealthiness.
Notably, FGSM excels over us on the IW-SSIM metric due to its emphasis on perturbing pixel values rather than image content. However, we outperform it significantly in attack performance and other stealthiness metrics.
We further provide qualitative experiments comparing our method with these methods in \cref{fig:compare}.
We consider deceiving both sub-tasks of ODs simultaneously as truly fooling the object detection model. The results in \cref{fig:compare} indicate that existing methods can only deceive either the regression or the classification task individually, \ie, causing the predicted  box to deviate or misclassifying. For instance, in the case of the LGP attack, it performs excellently in terms of attack metrics (\eg, achieving an mAP$_{50}$ of $1.49$ on Faster R-CNN across the MS-COCO dataset), but still fails to effectively deceive the regression sub-task, meaning the predicted boxes still remain on the main objects in the image.
The results from \cref{fig:compare} demonstrate that our approach outperforms others significantly, achieving true deception of the object detector, including prediction box deviations or disappearances and misclassification results.

\subsection{Ablation Study}
In this section, we explore the effect of different modules and backbones on our method.
We conduct experiments on the Faster R-CNN model with ResNet50 as the backbone across the MS-COCO dataset.

\noindent\textbf{The effect of different modules.} 
We investigate the effect of different modules on NumbOD. 
We use A, B, C, and D to represent $\mathcal{J}_{loc}$, $\mathcal{J}_{cls}$, $\mathcal{J}_{lfc}$, and $\mathcal{J}_{hfc}$, respectively.
Experimental results in \cref{fig:ablation_results} (a) demonstrate that none of the variants of our proposed method can match the performance of the complete version.

\noindent\textbf{The effect of backbone.} 
We examine the effect of different backbones on NumbOD, using three Faster R-CNN variants: \textit{ResNet50} (R50), \textit{ResNet101} (R101), and \textit{ResNeXt101} (X101). 
These models are tested on MS-COCO and PASCAL VOC datasets to evaluate NumbOD's attack performance. 
The results in  \cref{fig:ablation_results} (b) demonstrate our method's outstanding attack performance across various backbones.

\section{Defense}

\subsection{Corruption}
Corruption is a representative image preprocessing method used to mitigate adversarial examples. 
We employ two popular strategies, \textit{Brightness} (``B-'') and \textit{Spatter} (``S-''), to corrupt adversarial examples.
As illustrated in ~\cref{fig:defense} (a), the mAP$_{50}$ of the Faster R-CNN model decreases as the degree of corruption increases. 
However, our attack remains effective even when the erosion level is $5$, with an average mAP$_{50}$ value below $25\%$.
These findings indicate that NumbOD can effectively resist the corruption-based pre-processing defense.

\vspace{-0.2cm}
\subsection{Pruning \& Fine-tuning}
Pruning~\cite{zhu2017prune} involves selectively removing specific architectural components or parameters susceptible to exploitation by adversaries, thereby enhancing the resilience against adversarial attacks. As shown in ~\cref{fig:defense} (b), we select pruning rates from $0$ to $0.8$, demonstrating NumbOD's consistent ability to execute potent attacks even as the detector approaches collapse.
Similar to pruning, fine-tuning~\cite{peng2022fingerprinting} involves modifying the model to adjust inherited pre-trained weights. We conduct fine-tuning on five widely used object detectors to defend against adversarial examples. The results in ~\cref{fig:defense} (c) indicate an increase in the model's mAP$_{50}$ after fine-tuning, but NumbOD still maintains high attack performance.

\vspace{-0.2cm}
\subsection{Adversarial Training}
Adversarial training~\cite{PGD} is considered one of the most effective defense mechanisms against adversarial attacks, enhancing the robustness of models by introducing noise into the training dataset.
We fine-tune five well-trained object detectors from the MMDetection repository on the MS-COCO dataset. As shown in ~\cref{fig:defense} (d), our method maintains strong attack performance, with only a slight mAP$_{50}$ drop of less than $2.5\%$, even after adversarial training. This confirms our method's resilience against adversarial training.
 \vspace{-0.2cm}
\section{Conclusion}
In this paper, we propose NumbOD, the first 
model-agnostic spatial-frequency fusion attack against object detectors, rendering them numb to input images and unable to detect objects.
It consists of a spatial coordinated deviation attack and a critical frequency interference attack. 
We first design a dual-track attack target selection strategy, selecting the top-k high-quality bounding boxes independently from both classification and regression subtasks as attack targets.
Subsequently, we utilize directional induction to shift the detected bounding boxes output by the object detectors and devise a foreground-background separation attack to disrupt classification, thereby deceiving the model in the spatial domain. Concurrently, we distort the high-frequency information of images in the frequency domain to enhance the attack efficiency for critical objects.
Our extensive experiments on nine object detectors and two datasets show that our NumbOD achieves high attack performance and stealthiness, surpassing SOTA attacks against object detectors.

\vspace{-0.2cm}
\section*{Acknowledgements}
This work is supported by the National Natural Science Foundation of China (Grant No.62072204) and the National Key Research and Development Program of China (Grant No.2022YFB4502001). 
The computation is completed in the HPC Platform of Huazhong University of Science and Technology.
Dezhong Yao and Wei Wan are co-corresponding authors.

\bibliography{main}

\end{document}